\documentclass[12pt, twoside, a4paper]{article}
\usepackage{imm17_4}

\usepackage{color, colortbl}

\hypersetup{pdftitle=Implementation of deep learning algorithm for automatic detection of brain tumors using intraoperative IR-thermal mapping data}
\hypersetup{pdfauthor={A.V.Makarenko, Constructive Cybernetics Research Group}}
\hypersetup{pdfsubject={Machine Learning}}
\hypersetup{pdfkeywords={deep learning, brain tumors, IR-thermal mapping, cold probe, intraoperative navigation.}}

\newcommand{\MaincltStr}{arXiv.org}

\newcommand{\Leftclt}{A.\,V.\,Makarenko, M.\,G.\,Volovik}

\newcommand{\Rightclt}{Implementation of deep learning algorithm for automatic detection of brain tumors\ldots}

\newcommand{\UDK}{
PACS: 42.30.Sy, 42.62.Be, 87.19.La, 87.57.Nk, 87.57.nm.}

\begin{document}

\Mainclt % Печать колонтитула на первой странице

% Название статьи
\begin{center}
\Large{\bf Implementation of deep learning algorithm \\for automatic detection of brain tumors \\using intraoperative IR-thermal mapping data}\\[2ex]
\end{center}

% И.О.Ф автора и E-mail
\begin{center}
\large\bf{A.\,V.\,Makarenko }\supit{a,}\supit{b}
\footnote{E-mail: avm.science@mail.ru},
\large\bf{M.\,G.\,Volovik }\supit{c,}\supit{d}\\[2ex]
\end{center}

% Место работы автора
\begin{center}
\supit{a}\normalsize{Constructive Cybernetics Research Group}
\\
\normalsize{P.O.Box~560, Moscow, 101000 Russia}\\[3ex]

\supit{b}
\normalsize{Institute of Control Sciences, Russian Academy of Sciences}
\\
\normalsize{ul.~Profsoyuznaya~65, Moscow, 117977 Russia}\\[3ex]

\supit{c}\normalsize{Volga Federal Medical Research Center, Ministry of Health of Russia}
\\
\normalsize{Nizhny Novgorod, Russia}\\[3ex]

\supit{d}
\normalsize{Nizhny Novgorod State Medical Academy}
\\
\normalsize{Nizhny Novgorod, Russia}\\[3ex]
\end{center}

%\small{Received October 28, 2015}
% Аннотация к статье
\begin{quote}\small
{\bf Abstract}. The efficiency of deep machine learning for automatic delineation of tumor areas has been demonstrated for intraoperative neuronavigation using active IR-mapping with the use of the cold test. The proposed approach employs a matrix IR-imager to remotely register the space-time distribution of surface temperature pattern, which is determined by the dynamics of local cerebral blood flow. The advantages of this technique are non-invasiveness, zero risks for the health of patients and medical staff, low implementation and operational costs, ease and speed of use. Traditional IR-diagnostic technique has a crucial limitation - it involves a diagnostician who determines the boundaries of tumor areas, which gives rise to considerable uncertainty, which can lead to diagnosis errors that are difficult to control. The current study demonstrates that implementing deep learning algorithms allows to eliminate the explained drawback.
\end{quote}

% Ключевые слова - используем окружение Keyword*.

\begin{Keyworden}
deep learning, brain tumors, IR-thermal mapping, cold probe, intraoperative navigation.
\end{Keyworden}

% По умолчанию двойная нумерация формул,
% одинарная для  окружений,
% нужна иная --- переопределите в преамбуле.
% Следите за счётчиками самостоятельно!

% В начале каждой главы (если необходимо)
\setcounter{equation}{0}
\setcounter{lem}{0}
\setcounter{teo}{0}

% Обязательна автоматическая нумерация и автоматические
% ссылки на формулы и литературу.
% На все нумерованные формулы должны быть ссылки в тексте (ПОЖАЛУЙСТА, проверяйте!)

% Ссылки на литературу и формулы в
% теоремах, леммах и др. окружениях
% должны заключаться в символы $, например, $(\ref{lem2.3})$,
% чтобы они печатались прямым шрифтом.
% Для ссылок на формулы можно также использовать \eqref{2.3}.
% Если в окружениях встречаются знаки ; или : или скобки, они оформляются
% как \rm ({\rm(} {\rm;}) и т.п.

\section{Introduction}
\label{sec:intro}

Today in brain tumor removal, neurosurgery clinics use different techniques of intraoperative neuronavigation: CT, MRT, tractography, fluorescence diagnostics, laser biospectroscopy, etc. These technologies yield clinical data of various resolution, can be quite expensive and difficult to use, and have other limitations.

An alternative non-invasive intraoperative neuronavi\-gation technique is active IR-thermal mapping with the use of the cold probe~\cite{bib:article_Makarenko_JourOptTech_2015_82_459}. This approach employs a matrix IR-imager to remotely register the space-time distribution of surface temperature pattern, which is directly correlated with the dynamics of local cerebral blood flow. It has been established~\cite{bib:article_Pillaia_AJNR_2015_36_7} that in brain tumor invasion areas the topography of blood vessels is changed and the autoregulation of cerebral blood flow is disrupted.

As we have demonstrated~\cite{bib:article_Makarenko_JourOptTech_2015_82_467}, the isostatic distribution of the cortex temperature is not informative for the delineation of the pathology area. This approach of using IR-thermal mapping in diagnosis often yields negative research results, whereas the cold test, which disrupts the compensatory autoregulation mechanisms temporarily and reversibly, allows to make a successful functional and topical diagnosis~\cite{bib:article_Makarenko_JourOptTech_2015_82_467}. As a rule, this technique involves a visual assessment of IR-thermal maps by a trained specialist who determines the presence of a pathology area and its most probable boundaries. The human factor gives rise to considerable uncertainty, which can lead to diagnosis errors that are difficult to control. Our research demonstrates that implementing the algorithms of deep learning removes this limitation by automating the detection of pathology areas and delineation of the tumor projection on the cortex surface. The accuracy of our approach can potentially be developed to the degree where it can be effectively used in intraoperative planning of transcortical interventions for tumor removal by the surgeons.

\section{Active IR-thermal mapping technique}

The central feature of active IR-thermal mapping of the exposed area of dura mater and brain cortex before the removal of intracranial tumors is the application of the {\em cold probe}, i.e. uniform irrigation of dura mater and brain cortex in the trephine opening with NaCl~0.9\% solution at 20-22°С for 30~sec (RU~patent 2269287, 2006). The test perturbs the thermal regulation system of the brain to cause a perceptible variation in the dynamics of temperature pattern recovery within the tumor projection on the cortex surface and outside it (in the relatively intact cortex areas)~\cite{bib:article_Makarenko_JourOptTech_2015_82_459}.

\begin{figure}[!htb]
\begin{center}
\includegraphics[width=148mm, height=35mm]{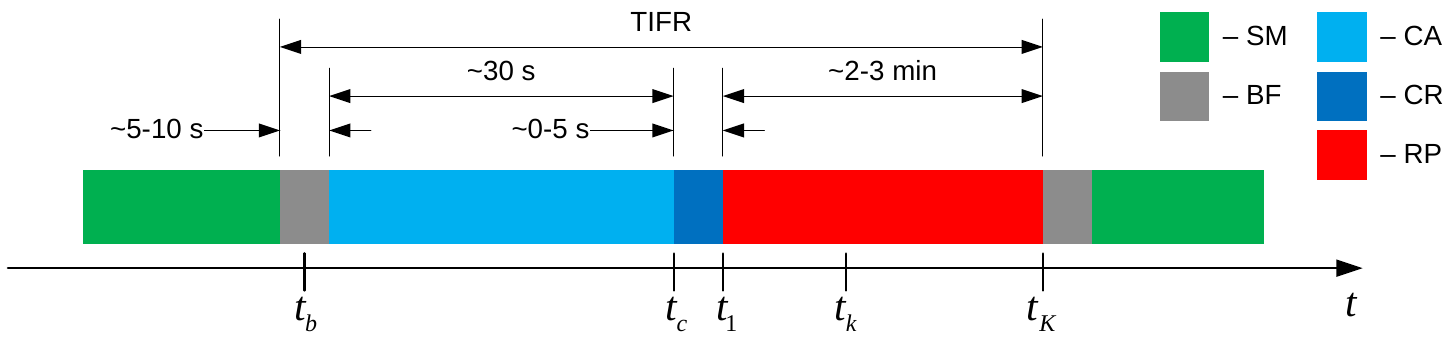}
\caption{IR-Monitoring Cyclogram (without timebase): TIFR -- thermal film recording; SM -- surgical manipulations; BF -- buffer time span; CA -- coolant application; CR -- passive or active (aspiration) removal of coolant; RP -- free temperature pattern recovery.}
\label{fig:cyclo_diag}
\end{center}
\end{figure}

After the craniotomy, we register the initial thermal map of the dura mater and after opening the dura mater -- a thermal map of the exposed cortex. A cold probe is applied at each of these stages. A cyclogram of IR-monitoring in relation to major procedures is shown in Fig.~\ref{fig:cyclo_diag} (for more detail see~\cite{bib:article_Makarenko_JourOptTech_2015_82_459}).

The readings are taken with a Thermo Tracer TH-9100 (NEC, Japan) IR-imager (Spectral Range: from 8~to 14~um; sensitivity: 0.025-0.03°С $±2$\% of reading; IR-Image Resolution 320~(H)~x~240~(V)~pixel). The IR-imager is mounted on a stand so that it is stationary and has the maximum view of the operational field when the viewing axis is perpendicular to it. This allows to avoid contamination or any interference with the surgical team manipulations. The distance between the lens and the cortex is~12 to 18~cm, which allows to use the capacity of the IR-imager matrix optimally. The object-side physical pixel size varies between 220-289~um.

\section{Automatic detection of brain tumors}

We classify the objects in the IR-imager field of view (frame Img) into the following area-types:
\begin{itemize}
\setlength{\parindent}{0pt}
\setlength{\parskip}{0.2ex plus 0.17ex minus 0.2ex}

\item WA -- working area, covering the tissue that is capable of actively responding to the cold test and where the thermal response is diagnostically significant;

\item NWA -- non-working area in the frame, complementary to WA area within the frame~Img;

\item DM -- dura mater over the brain cortex;

\item BC -- exposed cortex, complementary to~DM area within~WA;

\item NA -- relatively intact cortex (or dura mater, depending on the operation stage) within the~NA area;

\item HA -- tumor area (in projection on the cortex surface), complementary to~NA area within~WA.

\end{itemize}

There are additional secondary area-types:
\begin{equation*}
\begin{aligned}
\mathrm{NA\_BC} = \mathrm{NA} \cap\mathrm{BC}, \quad
\mathrm{NA\_DM} = \mathrm{NA} \cap\mathrm{DM}, \\
\mathrm{HA\_BC} = \mathrm{HA} \cap\mathrm{BC}, \quad
\mathrm{HA\_DM} = \mathrm{HA} \cap\mathrm{DM},
\end{aligned}
\end{equation*}
where $\cap$ is intersection of areas.

\begin{figure}[!htb]
\begin{center}
\includegraphics[width=136mm, height=44mm]{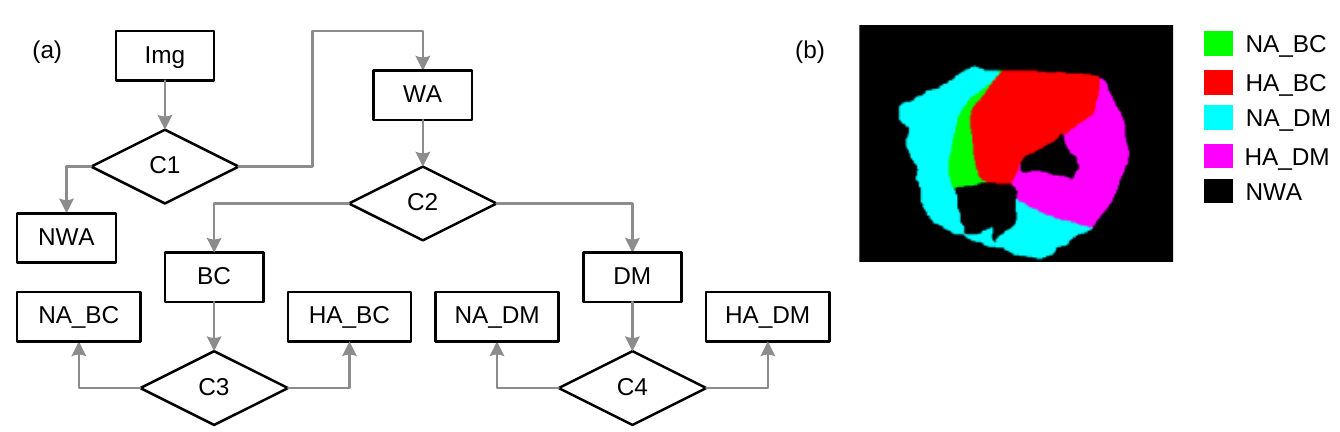}
\caption{(a) -- classification diagram, С1-С4 -- binary classifiers. (b) -- reference areas delineation within the frame (sample~CP52 from train/test sets).}
\label{fig:giag_class}
\end{center}
\end{figure}

The classification process (via binary classification) is shown in Fig.~\ref{fig:giag_class}(a), and a sample of reference areas delineation is shown in Fig.~\ref{fig:giag_class}(b).

With regard to the nature of neurosurgery and various tumor localizations (within the brain, on the cortex, or invading the dura mater) we devised three modes of classification (see Table~\ref{tbl:Regime_Class}): On -- WA area covers only dura mater and exposed cortex is positively absent from the frame; In -- WA area covers exposed cortex, as well as dura mater; Off -- WA area covers only exposed cortex and dura mater is positively absent from the frame.
\begin{table}[htb]
\renewcommand{\arraystretch}{1}
\newcolumntype{C}[1]{>{\hsize=#1\hsize\centering\arraybackslash}X}
\begin{center}
\caption{Classifier operation modes (* -- potentially present areas).}
\label{tbl:Regime_Class}
\small{
\begin{tabularx}{0.8\textwidth}[c]{C{0.5} C{0.5} C{0.5} C{0.5} C{0.5} C{0.5}}
\hline
Regime & NWA & NA\_BC & HA\_BC & NA\_DM & HA\_DM \\ \hline\hline
On     & *   &        &        & *      & * \\
In     & *   & *      & *      & *      & * \\
Off    & *   & *      & *      &        & \\ \hline
\end{tabularx}
}
\end{center}
\end{table}

\begin{figure}[htb]
\begin{center}
\includegraphics[width=167mm, height=37mm]{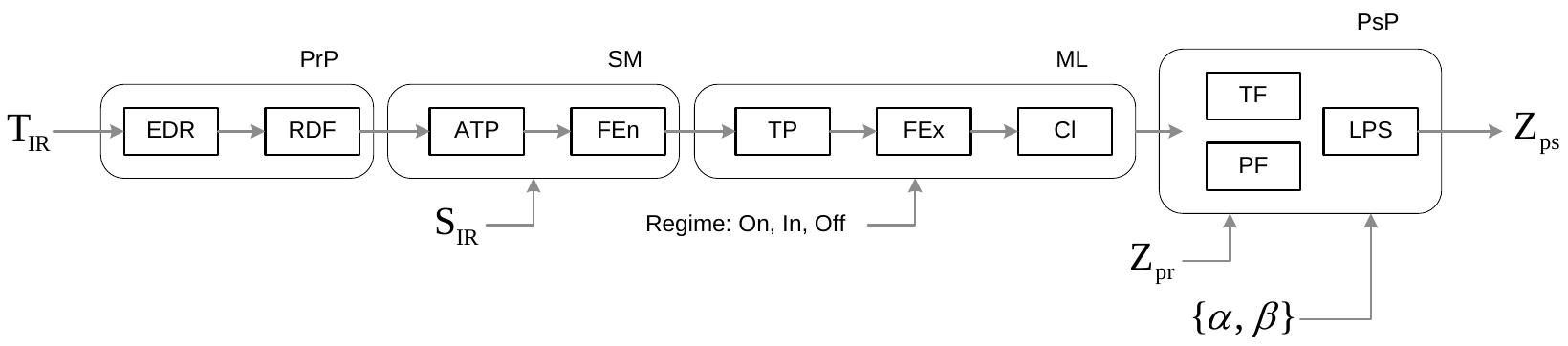}
\caption{Structural-functional diagram of area identification circuit in the IR-imager field of view: PrP -- pre-processing; SM -- system model; ML -- machine learning; PsP -- post-processing.}
\label{fig:giag_machine}
\end{center}
\end{figure}

We have proposed and positively tested the circuit for area identification within the field of view of the IR-imager (shown in Fig.~\ref{fig:giag_machine}). The system is fed TIR from the matrix IR-imager -- a 3D~array of pixel temperatures (2D -- spatial and~1D -- temporal components), which undergoes major processing: EDR -- eliminating minor frame shifts and/or rotation; RDF -- deleting damaged frames (due to fatal frame shifts and/or rotation or foreign objects in the frame); ATP -- approxi\-mation of temperature pixels; FEn -- features engineering; TP -- transformation of predictors; FEx -- feature hierarchy extraction; Cl -- classifier; TF -- topological filter; PF -- probabilistic filter; LPS -- logical-and-probabilistic solver.

The vector of parameters~$\mathrm{S_{IR}}$ governs the modeling of temperature pattern recovery dynamics. A priori area identification~$\mathrm{Z_{pr}}$ for delineating WA/NWA and BC/DM can be carried out in semiautomatic mode by analyzing TV channel data from the IR-imager using computer vision algorithms. Error levels of the second and first order~$\{\alpha,\,\beta\}$ set statistical thresholds for arriving at the final solution -- a posteriori area identification~$\mathrm{Z_{ps}}$.

The training and testing of the classifier involved the data of~71 active IR-thermal maps (from~43 neurosurgeri\-es) with operation mode counts at On -- 28, In -- 42, Off -- 1. Coolant application technique and data registration process were identical throughout. Pixel total amounted to~5\,452\,800: NWA -- 2\,781\,687; NA\_BC -- 288\,714; HA\_BC -- 419\,110; NA\_DM -- 1\,162\,225; HA\_DM -- 801\,064. The procedure of raw data acquisition and the technique of IR-frame mapping for compiling train and test subsets are described in detail in~\cite{bib:article_Makarenko_JourOptTech_2015_82_459}.

Research has shown that the data on~NA\_BC and~HA\_BC is insufficient for adequate learning of a stable classifier and at this stage we focus on studying the~On mode. We have carried out training of two types of classifiers: Random Forest (RF) and Stacked Denoising Auto-encoders (SDAE)~\cite{bib:article_Vincent_JourMachLearnRes_2010_11_3371}. For quality charac\-teristics of these models see Table~\ref{tbl:Param_Quality_Class}.
\begin{table}[ht!]
\renewcommand{\arraystretch}{1}
\newcolumntype{C}[1]{>{\hsize=#1\hsize\centering\arraybackslash}X}
\begin{center}
\caption{Per-pixel quality values of the learned classifier (Ac -- accuracy, Sn -- sensitivity), test set.}
\label{tbl:Param_Quality_Class}
\small{
\begin{tabularx}{0.8\textwidth}[c]{C{0.2} C{0.9} C{0.325} C{0.325} C{0.45}}\hline
Model & 95\% CI Ac       & Sn NA       & Sn HA      & Sn NWA \\ \hline\hline
RF    & (0.6942, 0.7022) & 0.7482      & 0.2812     & 0.7942 \\
SDAE  & (0.7158, 0.7235) & 0.7323      & 0.6120     & 0.7445 \\ \hline
\end{tabularx}
}
\end{center}
\end{table}

As Table~\ref{tbl:Param_Quality_Class} shows, the quality RF~classifier results in~HA area identification are inadequate, whereas SDAE~classifier yields nearly uniform results across the board and provides basic functionality. The study of the sources of such a difference between~RF and~SDAE models is the subject of our further research.

\section{Example}

\begin{figure}[t]
\begin{center}
\includegraphics[width=174mm, height=71mm]{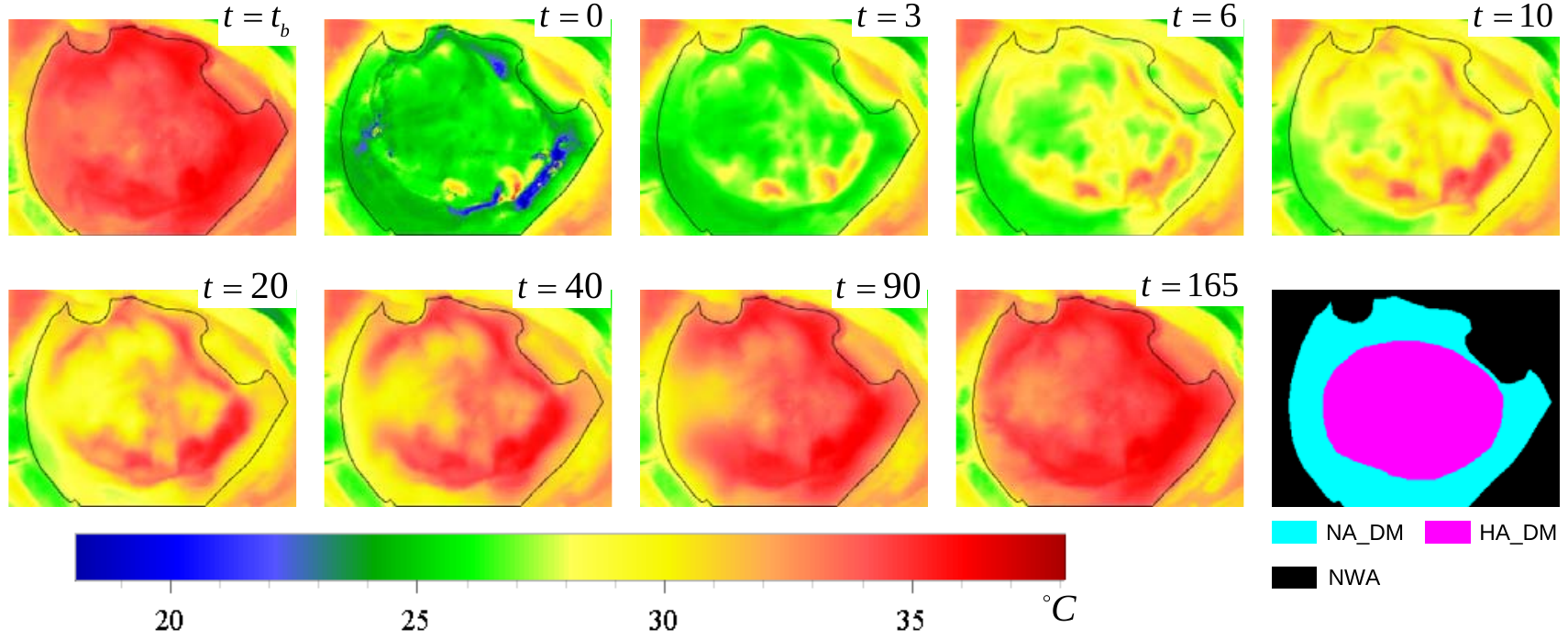}
\caption{IR-thermal maps sequence for sample~CP90. At time~$t=10$ the following projections are shown: a -- vessels connected to the tumor, b -- sinus.}
\label{fig:sample_CP_90_thermo}
\end{center}
\end{figure}

As an example of classifier operation, let us look at sample~CP90 (Grade~IV glioblastoma, localized in mid-posterior regions of left frontal lobe, parasagittally, subcortically, at~3-5~mm depth). This case was not included into train subset. Fig.~\ref{fig:sample_CP_90_thermo} shows an IR-frames sequence from an active IR-thermal mapping of unopened dura matter (classification mode -- On).

With sample~CP90 the algorithm yielded the following pixel by pixel quality values (Balanced Accuracy, classifier output): NA -- 68.11\%, HA -- 68.45\%, WA -- 94.55\%.

\begin{figure}[htb]
\begin{center}
\includegraphics[width=131mm, height=65mm]{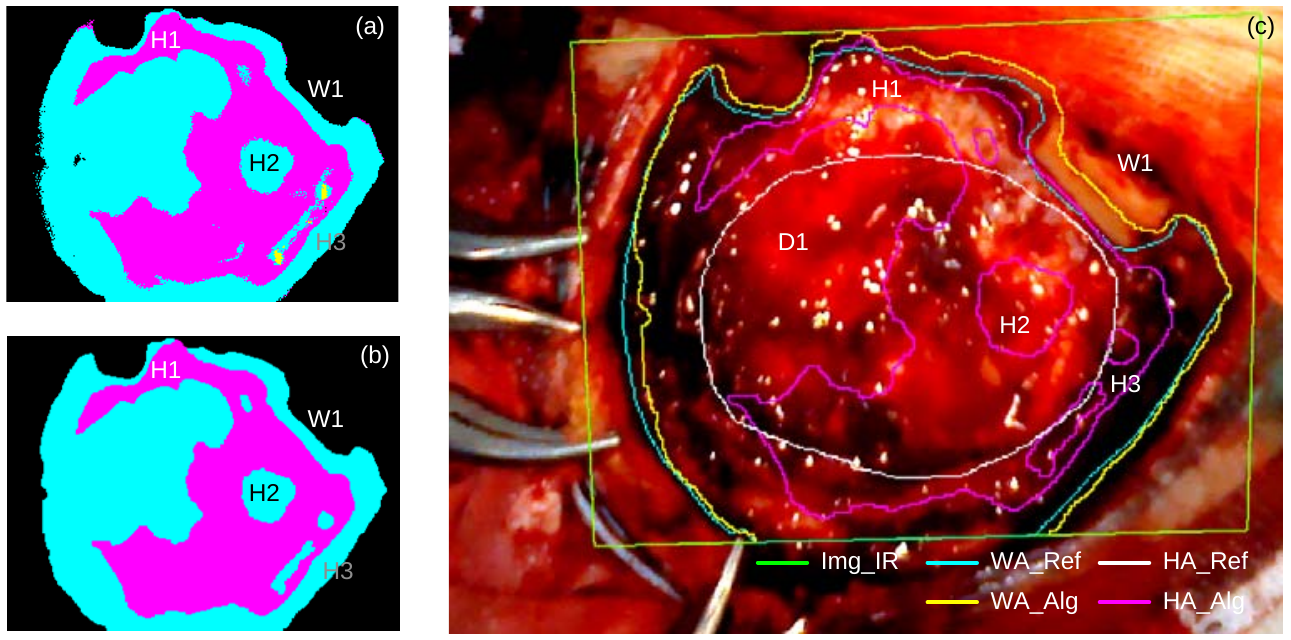}
\caption{Sample CP90. Output: (a) -- classifier, (b)~--~topological filter. (c)~--~IR-imager TV~channel (at time~$t_b$) overlaid with IR-thermal frame boundaries and reference areas: Img\_IR -- IR-frame boundary; WA\_Ref and~WA\_Alg -- WA/NWA~areas boundary by initial mapping and by classifier (after topological filter) respectively; HA\_Ref and~HA\_Alg -- NA/HA~areas boundary.}
\label{fig:sample_CP_90_Mask}
\end{center}
\end{figure}

As Fig~\ref{fig:sample_CP_90_Mask}(a) demonstrates, there is spot noise and areas of small physical size in the classifier output. These two types of errors are efficiently corrected by the topological filter, see Fig.~\ref{fig:sample_CP_90_Mask}(b). Beside these errors, there are also misidentified areas of considerable dimensions. The areas that require particular attention are (see Fig.~\ref{fig:sample_CP_90_Mask}(c)): W1 -- a segment of the scull bone (NWA area identified as WA area); H1 -- blood vessels connected to the tumor (NA$\to$HA); H3 -- the sinus that is partially invaded by the tumor and functions abnormally (NA$\to$HA).

\section{Conclusion}

The study, based on the processing of data from~43 neurosurgeries, shows that in principle automatic delineation of brain tumor areas with the use of active IR~thermal mapping data is possible. The quality characteristics of the trained classifier that we have achieved (see Table~\ref{tbl:Param_Quality_Class}) ensure basic functionality of the developed algorithm, however these characteristics are not final and should be considered a reference point for the further increase in accuracy of the system. In light of this, we are currently working on the following tasks.

First, the~SDAE classifier, that we implemented, is a fully connected multi-layer neural network and, due to its topology, it cannot fully make use of the potential hierarchical nature of the informative features (recovery and synthesis of the temporal structure of dynamics on various time scales). To remove this limitation we plan to implement a classifier based on 1D~CNN~\cite{bib:article_Martinez_CompIntelMagIEEE_2013_8_20}.

Second, the fully connected neural network makes a per-pixel decision based on the processing of the data of a single pixel (only for the temporal dynamics of the free recovery of the temperature pattern, see Fig.~\ref{fig:cyclo_diag}). Thus, the classifier completely ignores the spatial context (which is only indirectly recovered by the topological filter). Consequentially, the classifier cannot verify nor the arteries that supply blood to the tumor or surround the tumor node, neither the significant veins near the tumor that connect to the venous sinuses. However, this data is provided by active IR~thermal mapping (e.g. see Fig.~\ref{fig:sample_CP_90_thermo}). Probably, this is what causes the errors in~H1 and~H3 areas, see Fig~\ref{fig:sample_CP_90_Mask}(c). To remove this limitation we plan to implement~TDSN~\cite{bib:article_Hutchinson_TransPatAnalMachIntelIEEE_2013_35_1944} and~RCNN~\cite{bib:report_Liang_CVPR_2015_3367} approaches.

Third, in our plans to increase the volume of training and testing sets.

It stands to mention that we are studying the structure of the trained classifier in order to determine the formal properties that differentiate~NA and~HA areas. This research is valuable for assessing the diagnostic potential of the technology in question, as well as for increasing the quality of classification by targeted transformation of predictors.

% Библиография, используем окружение Biblio*
\begin{Biblioen}

\bibitem{bib:article_Makarenko_JourOptTech_2015_82_459}
{\it A.V.~Makarenko and M.G.~Volovik}, Method of differentiated analysis of IR thermal maps of the exposed cerebral cortex when neurosurgical operations are being performed, J.Opt.Technol. {\bf 82}:7, 459 (2015).

\bibitem{bib:article_Pillaia_AJNR_2015_36_7}
{\it J.J.~Pillaia and D.J.~Mikulisb}, Cerebrovascular Reactivity Mapping: An Evolving Standard for Clinical Functional Imaging, American Journal of Neuroradiology {\bf 36}:1, 7 (2015).

\bibitem{bib:article_Makarenko_JourOptTech_2015_82_467}
{\it M.G.~Volovik and A.V.~Makarenko}, Parameters of the thermal patterns of the exposed cortex from the results of IR thermal mapping when tumors are being removed from the human brain, J.Opt.Technol. {\bf 82}:7, 467 (2015).

\bibitem{bib:article_Vincent_JourMachLearnRes_2010_11_3371}
{\it P.~Vincent, H.~Larochelle, I.~Lajoie, Y.~Bengio, and P.-A.~Manzagol}, Stacked Denoising Autoencoders: Learning Useful Representations in a Deep Network with a Local Denoising Criterion, Journal of Machine Learning Research {\bf 11}:2, 3371 (2010).

\bibitem{bib:article_Martinez_CompIntelMagIEEE_2013_8_20}
{\it H.P.~Martinez, Y.~Bengio, and G.N.~Yannakakis}, Learning deep physiological models of affect, Computational Intelligence Magazine, IEEE {\bf 8}:2, 20 (2013).

\bibitem{bib:article_Hutchinson_TransPatAnalMachIntelIEEE_2013_35_1944}
{\it B.~Hutchinson, L.~Deng, and D.~Yu}, Tensor Deep Stacking Networks, IEEE Transactions on Pattern Analysis and Machine Intelligence {\bf 35}:8, 1944 (2013).

\bibitem{bib:report_Liang_CVPR_2015_3367}
{\it  M.~Liang and X.~Hu}, Recurrent Convolutional Neural Network for Object Recognition, The IEEE Conference on Computer Vision and Pattern Recognition (CVPR), IEEE, 3367--3375 (2015).

\end{Biblioen}

%Обязательна информация об авторе

\noindent
\\\textsf{\textbf{Andrey V. Makarenko} -- was born in~1977, since~2002 -- Ph.~D. of Cybernetics. Founder and leader of the Research \& Development group "Constructive Cybernetics". Author and coauthor of more than 60~scientific articles and reports. Member~IEEE (IEEE Signal Processing Society Membership; IEEE Computational Intelligence Society Membership). Research interests: Analysis of the structure dynamic processes, predictability; Detection, classification and diagnosis is not fully observed objects (patterns); Synchronization and self-organization in nonlinear and chaotic systems; System analysis and math.~modeling of economic, financial, social and bio-physical systems and processes; Convergence of Data~Science, Nonlinear~Dynamics and~Network-Centric.}

\end{document}